\documentclass[11pt]{article}

\usepackage{acl}

\usepackage{times}
\usepackage{latexsym}

\usepackage[T1]{fontenc}

\usepackage[utf8]{inputenc}

\usepackage{microtype}
\usepackage{booktabs}
\usepackage{makecell}
\usepackage{graphicx}
\usepackage{mathtools}
\usepackage{enumitem}
\usepackage{float}
\usepackage{colortbl}

\usepackage{pifont}
\usepackage{multirow}
\usepackage{amsmath}
\usepackage{amssymb}

\usepackage{hyperref}
\hypersetup{colorlinks,allcolors=black}

\newcommand{\cmark}{\textcolor{green}{\ding{51}}}%
\newcommand{\xmark}{\textcolor{red}{\ding{55}}}%

\DeclareMathOperator*{\argmax}{argmax}  

\newcommand{\RNum}[1]{\uppercase\expandafter{\romannumeral #1\relax}}

\newcommand{\task}{CuQA}

\newcommand{\ie}{\textit{i}.\textit{e}.}
\newcommand{\eg}{\textit{e}.\textit{g}.}

\newcommand\Tstrut{\rule{0pt}{2.6ex}}       
\newcommand\Bstrut{\rule[-0.9ex]{0pt}{0pt}} 
\newcommand{\TBstrut}{\Tstrut\Bstrut} 

\title{Plug-and-Play Adaptation for Continuously-updated QA}

\author{Kyungjae Lee$^{5}$ ~~~ \quad\quad Wookje Han$^1$ ~~~ \quad\quad Seung-won Hwang$^1$\thanks{~~correspond to seungwonh@snu.ac.kr} \\ \bf{Hwaran Lee$^2$ ~~~ \quad\quad Joonsuk Park$^{2,4}$ ~~~ \quad\quad Sang-Woo Lee$^{2,3}$} \\ ~~ \quad $^1$Seoul National University ~~~ \quad\quad $^2$NAVER AI Lab ~~~\quad\quad $^3$NAVER CLOVA ~~~~ \quad \\ $^4$University of Richmond \quad\quad\quad $^5$LG AI Research ~~~~ }

\begin{document}

\maketitle

\begin{abstract}
Language models (LMs) have shown great potential as implicit knowledge bases (KBs). And for their practical use, knowledge in LMs need to be updated periodically. However, existing tasks to assess LMs' efficacy as KBs do not adequately consider multiple large-scale updates. To this end, we first propose a novel task---Continuously-updated QA (CuQA)---in which multiple large-scale updates are made to LMs, and the performance is measured with respect to the success in adding and updating knowledge while retaining existing knowledge. We then present LMs with plug-in modules that effectively handle the updates. Experiments conducted on zsRE QA and NQ datasets show that our method outperforms existing approaches. We find that our method is 4x more effective in terms of updates/forgets ratio, compared to a fine-tuning baseline.

\end{abstract}

\section{Introduction}

LM-as-KB is a new paradigm in which pre-trained language models (LMs) are used as implicit knowledge bases (KBs)~\cite{petroni2019language}. This is made possible by LMs' impressive ability to memorize factual knowledge~\cite{heinzerling2021language,brown2020language}. Recently, two tasks have been used to assess such ability: LAMA, a knowledge probing benchmark, challenges LMs to fill in masked words over relational knowledge \cite{petroni2019language}; and closed-book QA (CBQA) examines whether LMs can correctly answer natural language questions \cite{roberts2020much}. 

\begin{figure}[t]
	\centering
	\includegraphics[width=74mm]{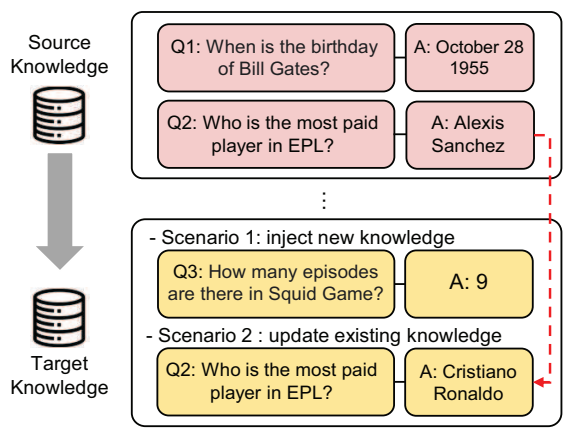}
	\caption{Examples of CuQA showcasing two scenarios.}
	\label{figure1}
\end{figure}

For practical usage, LM-as-KB requires that LMs are updated periodically to stay current with the ever-evolving world. 
Thus, LMs' ability to update knowledge should also be evaluated. 
To this end, we present \textbf{Continuously-updated QA (CuQA)}, which tests the ability to continuously inject knowledge to update (or \textbf{target knowledge}), while retaining existing knowledge (or \textbf{source knowledge}). 
Specifically, we consider multiple large-scale knowledge updates (8k to 60k) covering two scenarios: injecting new knowledge (Scenario 1 in Figure \ref{figure1}) and updating existing knowledge (Scenario 2 in Figure \ref{figure1}) . 

Our goal is to organize the implicit storage of knowledge, to add target knowledge (\textit{yellow box} in Figure \ref{figure1}) and anchor to select target knowledge.
A simple approach is to train updated LMs from scratch; however, this is far too expensive considering the parameter sizes of recent LMs, such as 175B for GPT-3 ~\cite{brown2020language} and about 11B for T5 \cite{raffel2020exploring}.
There has also been related work for the two scenarios. 
For Scenario 1, a method for continual learning can be adopted, constraining the distance between parameters before and after fine-tuning~\cite{chen2020recall}. 
However, this approach still suffers from so-called catastrophic forgetting, where the LMs fail to retain large amounts of source knowledge.
For Scenario 2, one may consider knowledge editing methods, where we see reasonable performances for a single knowledge edit while retaining the rest \cite{de2021editing,mitchell2021fast}. 
However, this line of work does not perform well when multiple edits are accumulated, \eg, only 67\% of 125 edits were updated, as reported in~\cite{mitchell2021fast}.

We propose to efficiently extend LMs with plug-and-play modules that store target knowledge. 
More specifically, we adopt a parameter-expansion method in which the LM storing existing knowledge is extended with plug-in feed-forward modules storing updated knowledge. 
Depending on the input, the LM selectively uses either the original LM or a plug-in module. 
We stress that, by keeping the original LM intact, we retain (a) not only source knowledge, (b) but also those outdated from updates (\textit{red arrow} in Figure \ref{figure1}). (a) is important to avoid catastrophic forgetting, while (b) is useful when updates need to be reverted due to ethical concerns---for example, there can be malicious attempts to override facts.

We evaluate our approach on zsRE~\cite{levy2017zero} and Natural Questions~\cite{kwiatkowski2019natural} to showcase
successful updates of new knowledge and retention of existing knowledge. 
We measure the accuracies on both previous and updated knowledge and find that ours show x4 higher updates/forgets ratio, compared to fine-tuning.
We will release our code and dataset.

Our key contributions are as follows:
\vspace{-2mm}
\begin{itemize}[leftmargin=0.4cm]
\setlength\itemsep{2pt}
    \item We present \task, a novel task to assess LMs' ability to continuously inject knowledge to update.
    \vspace{-2mm}
    \item We propose a new methodology, plug-and-play adaptation, to continually learn new knowledge while better retaining existing knowledge.
\end{itemize}
\section{Related Work}


The relevant research can be categorized into three groups: Knowledge Editing, Continual Learning, and Adaptation.
In Table \ref{table1}, we compare these with our method.

\paragraph{Editing Implicit Knowledge}
In Table \ref{table1}(a), knowledge editing methods~\cite{de2021editing,mitchell2021fast,dai2021knowledge}
aim to efficiently edit model's parameters on examples that have conflicts with old facts,
while preserving the outputs of untargeted examples.
Instead of directly updating gradients by fine-tuning, these methods transform the gradients for new edit parameters.
As representative methods for knowledge editing, KnowledgeEditor (KE) \cite{de2021editing} using LSTM produces gate vectors, then the gated sum of gradients is updated into the model, while MEND \cite{mitchell2021fast} uses simple MLP layers and residual connections for the same purpose.
Although these methods succeeded in updating the target examples less forgetting, their target scenario is
a single edit, such that the cumulative effect of multiple edits does not reflect well, which
disqualifies its use for our target task of update large-scale data (8K$\sim$60K).
As reported in \cite{mitchell2021fast}, MEND successfully updates only 67\% of edits when applying 125 edits, 
while our finding was consistent when none of the 125 edits was applied in our evaluation.\footnote{In the case of KE, we reimplement the released code for testing: https://github.com/nicola-decao/KnowledgeEditor.}
In addition, for editing previous knowledge, KE and MEND simulate knowledge updates, by
generating synthetic knowledge from LM.
Such generations may not be realistic data and also give unfair advantages to LM-based methods, while we use actual up-to-date knowledge as new data, which 
were annotated on recent corpus~\cite{zhang2021situatedqa}.

\begin{table}[t]
\centering
\scalebox{1.0}{
\resizebox{\columnwidth}{!}{
\begin{tabular}{lccc}
\noalign{\hrule height 1pt} 
~~ Method & \makecell{Forgetting \\ less?} & \makecell{Scales to\\a large set?} & \makecell{Conflict with\\old facts?}\\ 
\noalign{\hrule height 0.5pt} 
\TBstrut
(a) Editing & \cmark & \xmark  & \cmark \TBstrut \\
(b) CL & \cmark & \cmark& \xmark \TBstrut \\
(c) Adaptation & \xmark & \cmark & \xmark  \TBstrut \\ \hline
Our Method & \cmark & \cmark & \cmark \TBstrut \\
\noalign{\hrule height 1pt} 
\end{tabular}}}
\caption{Conceptual comparison of existing approaches.}
\label{table1}
\end{table}


\paragraph{Continual Learning (CL) for NLP}
For our task, we can adopt CL methods, learning a new task while preserving the accuracy on previous tasks.
\citet{kirkpatrick2017overcoming} proposed Elastic Weight Consolidation, 
alleviating catastrophic forgetting.
This method regularizes learning on a new task, by constraining the parameters trained on the previous task.
For NLP tasks, RecAdam \cite{chen2020recall} uses the regularization and annealing technique, which is a CL baseline in our experiment.
While CL approaches focusing on forgetting do not consider conflicts between old and new knowledge, our work deals with such a realistic scenario.
Additionally, previous work~\cite{dhingra2021time} proposed benchmarks for probing temporal language models, 
asking ``Fill-in-the-Blank (FIB)" questions.
Meanwhile, FIB questions are limited to evaluate masked language models, such as BERT and RoBERTa.
We extend to evaluate arbitrary questions for a knowledge-intensive task; closed-book QA, which can evaluate generative LMs with broader applicability, to include T5 and GPT.


\paragraph{Task-aware Adaptation for Transformers}
Recent works~\cite{hu2021lora,wang2020k,lin2020exploring} study LM adaptation to new labeled data in a new domain, which has a different data distribution from that at pretraining.
These works show performance improvements on downstream tasks in the new domain, while fine-tuning a small number of parameters. 
However, these adaptation methods do not consider sequential training, and overwrite the new data into the parameters that store previous knowledge.
In our experiment, it is observed that the adaptation methods are rapidly forgetting previously seen data, while performing well on new knowledge.


\section{A Continuously-updated QA Task}
\label{sec:task}
\paragraph{Task Description}


In this section, we propose Continuously-updated QA (\task), a new continual learning task for knowledge updates in LMs based on closed-book QA (CBQA)~\cite{roberts2020much}.
In CBQA, LMs answer factual questions with the implicit knowledge stored in the model, without any external context (\ie, in contrast to open-domain QA), so that LMs are required to adequately update their parameters to the target knowledge.
In our \task, LMs learn source (original) knowledge first, then update them with target (new) knowledge without source knowledge access.
For the above setting, source knowledge (to be retained) and target knowledge (to be added) in CuQA do not have any overlap of QA pairs (or paraphrases) for any given fact.


Specifically, we denote a factual pair of question and answer as $(q,a)$, source knowledge as $\mathcal{K}_s$, and target as $\mathcal{K}_t$.
We first build an initial model $\theta^{old}$ pre-trained on source knowledge $\mathcal{K}_s$. 
Then, we inject target knowledge $\mathcal{K}_t$ into the pre-trained model and obtain the infused model $\theta^{new}$.
Our goal is to memorize $\mathcal{K}_t$ on model $\theta^{new}$, with less forgetting $\mathcal{K}_s$.
If knowledge in $\mathcal{K}_t$ conflicts one in $\mathcal{K}_s$, 
the model is required to adjust its parameters by reflecting the target knowledge.
Note that multiple target knowledge can be sequentially updated to the model (see details in Section 4).

\paragraph{Research Questions}
CuQA is designed to address the following research questions:
\begin{itemize}[leftmargin=0.4cm]
\setlength\itemsep{2pt}
    \item RQ1: Can the method learn target knowledge while retaining source knowledge?
    \item RQ2: 
    How does sequentially learning multiple target knowledge affect the performance?
    \item RQ3: 
    How does the size of each target knowledge affect the performance?
\end{itemize}

\paragraph{Metric}
For evaluation, we measure the success of updates, retaining of source knowledge, and generality using exact match (EM) scores.
Additionally, we measure the ratio of forgets to updates.
\begin{itemize}[leftmargin=0.4cm]
\setlength\itemsep{2pt}
    \item \textbf{Accuracy on $\mathcal{K}_t$} : we evaluate how much model $\theta^{new}$ \textbf{successfully updates} examples in $\mathcal{K}_t$. 
    \item \textbf{Accuracy on $\mathcal{K}_s$}: how much model $\theta^{new}$ \textbf{forgets} examples in $\mathcal{K}_s$. This indicates performance degradation, when replacing $\theta^{old}$ with $\theta^{new}$.
    \item \textbf{Accuracy on $\mathcal{P}_s$, $\mathcal{P}_t$}: how well model $\theta^{new}$ \textbf{generalizes} on semantically equivalent questions (or paraphrases).
    \item \textbf{F/U Ratio} ($\#$ of forgets$\slash\#$ of updates): how many examples in $\mathcal{K}_s$ are forgotten per an update of one example in $\mathcal{K}_t$. ($\#$ of forgets) is equal to the difference of correct prediction cases in $\mathcal{K}_s$, between $\theta^{old}$ and $\theta^{new}$.
    
\end{itemize}

\section{Method}
In this section, we describe baseline approaches (Section \ref{ssec:baseline}), and introduce our proposed method for plug-and-play adaptation (Section \ref{ssec:ours}).

\begin{figure*}[t]
	\centering
	\includegraphics[width=156mm]{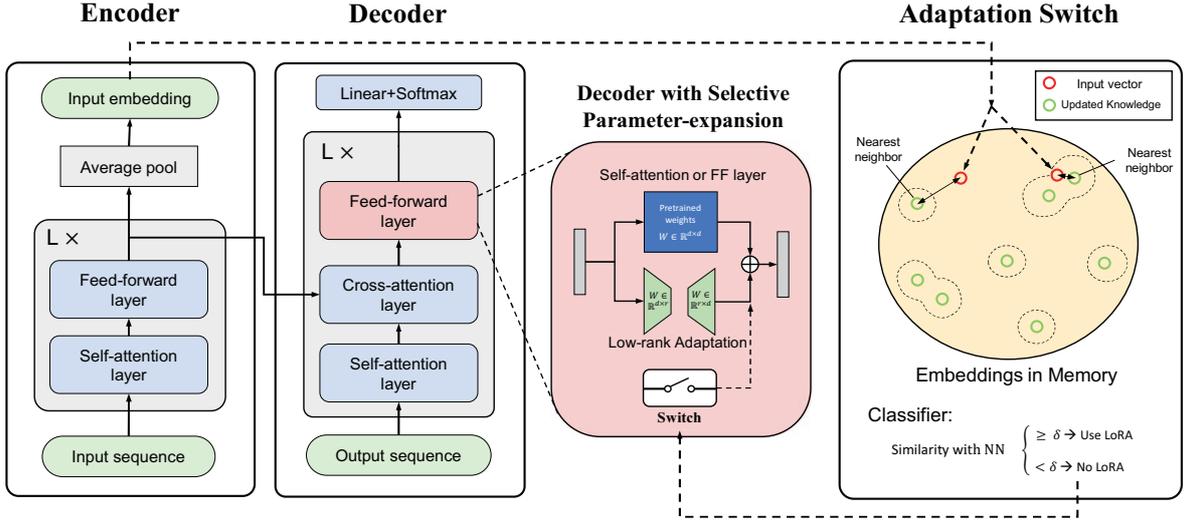}
	\caption{An overview of our proposed architecture.}
	\label{figure2}
\end{figure*}

\subsection{Baseline Approaches}
\label{ssec:baseline}

We establish three baseline for (a), (b), and (c), in Table \ref{table1}.
Since we found that a knowledge editing approach is outperformed by fine-tuning, we exclude it as baselines, and add fine-tuning instead.

\paragraph{Fine-tuning on target knowledge}

As a naive baseline, we start with the previous work~\cite{roberts2020much} for CBQA, by fine-tuning T5~\cite{raffel2020exploring} with encoder-decoder structure.
This baseline is to fine-tune the pre-trained model $\theta^{old}$ on facts in $\mathcal{K}_t$ to minimize the loss:
\begin{equation}
    \label{fine-tuning}
    \mathcal{L}_{FT} = \sum_{(q,a) \in \mathcal{K}_t} L((q,a) ; \theta)
\end{equation}
where $L$ refers to a seq2seq loss.
This baseline is expected to optimize accuracy on target knowledge $\mathcal{K}_t$, thus increases the distance between the before- ($\theta^{old}$) and after-parameters ($\theta^{new}$) resulting in the risk of forgetting.
For other baselines and our method, we adopt the same transformer: T5 as backbone network.

\paragraph{Regularized fine-tuning for CL}
We adopt RecAdam~\cite{chen2020recall} aiming to reduce the forgetting risk by adding a constraint to minimize the distance between $\theta^{old}$ and $\theta^{new}$ as follow:
\begin{equation}
    \label{constraint}
    \mathcal{R} =  \|(\theta - \theta^{old}) \|_p
\end{equation}
where $\| \cdot  \|_p$ indicates $L_p$ norm.
In addition, RecAdam uses an annealing technique, controlling the ratio between $\mathcal{R}$ and the fine-tuning loss (Eq. (\ref{fine-tuning})) as follows:
\begin{align}
    \mathcal{L}_{total} =&  ~ \lambda(t) \mathcal{L}_{FT}  + (1-\lambda(t)) \mathcal{R}, \\
    \lambda(t) =& ~ \frac{1}{1+\textrm{exp}(-k \cdot (t-t_0))}
\end{align}
where $k$ and $t_0$ are hyper-parameters.


\paragraph{Adapters for knowledge updates}
For adaptation approaches, we implement two parameter-expansion methods: K-adapter \cite{wang2020k} and LoRA \cite{hu2021lora}.
The approaches freeze the parameters $\theta^{old}$ in pre-trained LM and augment additional new parameters $\tilde{\theta}$ in the LM to train target knowledge as following:
\begin{equation}
    \mathcal{L}_{adap} = \sum_{(q,a) \in \mathcal{K}_t} L((q,a) ; 
    \theta^{old},
    \tilde{\theta}).
\end{equation}
For $\tilde{\theta}$, K-adapter \cite{wang2020k} uses augmented self-attention layers, while LoRA \cite{hu2021lora} utilizes extra low-rank matrices.

\subsection{Our Method}
\label{ssec:ours}

Motivated by the intuition of regularization to preserve source knowledge and that of adapters to inject target knowledge into new parameters, we show their strengths can be combined for our task.
At the inference phase, our method selectively uses the plug-in modules to keep source knowledge intact, while tasks requiring target knowledge will be redirected to new plug-in modules.

%



Specifically, our distinction is augmenting function $f$ (in an original LM) with function $g$, representing source and target knowledge respectively.
The function $f$ is a single layer in transformer trained on source knowledge $\mathcal{K}_s$, and $g$ is an augmented function with new parameters for $\mathcal{K}_t$.
Existing work, such as LoRA, can be interpreted by adding the two functions:
\begin{equation}
    \label{eq_lora1}
    h = f(x) + g(x)
\end{equation}
where $f$ is one-linear layer in self-attention or feed-forward layers.
That is, $f(x)=W_0 x$, where $W_0 \in \mathrm{R}^{d \times k}$ denotes the pre-trained and fixed parameters.
LoRA uses low-rank matrices as $g(x)$, \ie, $g(x) = BAx$, where $B \in \mathrm{R}^{d \times r}$, $A \in \mathrm{R}^{r \times k}$, and  $r <\!\!<  min(d,k)$.
The low-rank matrices $A$ and $B$ are trainable parameters for updating target knowledge.
The new layer with the additional matrices is denoted as follows:
\begin{equation}
    \label{eq_lora}
    h = W_0x + BAx = (W_0 + BA)x
\end{equation}
However, the above add-aggregation has a limitation, as $g(x)$ can affect the model's outputs, and increase the distance between hidden states in $\theta^{old}$ and $\theta^{new}$, which causes a forgetting problem.



Our key distinction is adding a selector, that is selectively activated for $q$ requiring the use of plug-in module $g$, as follows:
\begin{equation}
    \label{aggregation}
    h = f(x) + \sigma(q) \cdot g(x)  
\end{equation}
where $\sigma(q)$ is 1 or 0 depending on query $q$.
While there can be various ways to train the selector in a sophisticated way, supervised either directly, or indirectly in an end-to-end manner, 
we show a simple unsupervised selector is already sufficient to show gains. Specifically, our selector is a key-value lookup where the key is $m_i$ and value is $g$.
At inference time, when given query $q$ is based on facts in $\mathcal{K}_t$, we activate the augmented $g$ for generating its output.
If $q$ is not from $\mathcal{K}_t$, we use only the original model $\theta^{old}$ for generation.
To classify whether the input is from $\mathcal{K}_t$ or not, 
we build explicit memory with embeddings of $\mathcal{K}_t$ and leverage the distance with nearest neighbor (NN) in the memory.


Let $\mathcal{M} \in \mathrm{R}^{N \times d}$ be memory embeddings that stores embeddings of input questions in $\mathcal{K}_t$, where $N$ is the total number of examples in $\mathcal{K}_t$.
As shown in Figure \ref{figure2}, question embedding can be extracted from the encoder, by averaging the hidden states of input sequence.
In T5 model with encoder-decoder, this averaging method is known to be effective on semantic textual similarity, as in \cite{ni2021sentence}.
Given question $q$, cosine similarity with NN is calculated as follows:
\begin{equation}
    s_q = \textrm{max}_i ( \textrm{sim} (m_i, q) ), ~~  m_i \in \mathcal{M}
\end{equation}
where $\textrm{sim}$ indicates cosine similarity.
Based on $s_q$, if the score is greater than or equal to threshold $\delta$, we assume $q$ is from target knowledge $\mathcal{K}_t$.
We build a indicator function as follows:
\begin{equation}
    \sigma (q)= 
    \begin{cases}
        ~~ 1 ~~~~~ \textrm{if} ~~ s_q \geq \delta, \\
        ~~ 0 ~~~~~ \textrm{if} ~~ s_q < \delta. \\
    \end{cases}
\end{equation}
In other words, $s_q \geq \delta$ indicates that input $q$ is semantically similar with one fact in $\mathcal{K}_t$. 
At that time, our model is augmented with $g$ that stores new and updated knowledge.

Meanwhile, as shown in Figure \ref{figure2}, we apply the selective use of parameters to only a decoder in a transformer architecture, not a encoder.
The switch $\sigma$ depends on query embedding $q$, and the embedding $q$ is extracted from T5 encoder.
If we apply the switch $\sigma$ to hidden states in T5 encoder, this causes a recursion relation, or inefficient computations.
By augmenting $g$ for the decoder, embedding $q$ is not changing during updating target knowledge, and depends on only pre-trained $\theta^{old}$.

\paragraph{General case of multiple knowledge updates}
Our new perspective has another benefit of naturally generalizing to sequential ($>$2) sources.
Assume that there are multiple target knowledge to be sequentially updated, \ie, $\mathcal{K}_t^1, \mathcal{K}_t^2,...,\mathcal{K}_t^M$.
We build multiple functions $g_k$ and memories $\mathcal{M}_k$ (where $k=1,...M$), according to each target knowledge.
The new function considering the multiple knowledge is denoted as follows:
\begin{equation}
    \label{multiple}
    h = f(x) + \sum_{k=1}^{M} \sigma_k(q) \cdot g_k(x)
\end{equation}
During training $j$-th target $\mathcal{K}_t^j$, the switch $\sigma_k(q)$ is activated where $1 \leq k \leq j$.
At inference time, our selector extracts top1-NN fact $m^*$, which is closest to a query $q$.
If $m^*$ is in $\mathcal{M}_k$, the switch $\sigma_j(q)$ is activated where $1 \leq j \leq k$, as follows:
\begin{equation}
    m^* =   \argmax_m (\textrm{sim} (m, q)), ~~ m \in \mathcal{M}_{1:M}  \\
\end{equation}
If the NN fact $m^*$ is in $\mathcal{M}_j$, we estimate that its implicit knowledge is stored in the accumulated function $\sum_{k=1}^{j} g_k(x)$.
That is, when $m^*$ is in $\mathcal{M}_j$, the activation is decided as follows:
\begin{equation}
    \label{multiple_switch}
    \begin{split}
        \sigma_k (q) = &
        \begin{cases}
            ~~ 1 ~~~~~ \textrm{if} ~~ s_q \geq \delta ~~ \textrm{and} ~~ 1 \leq k \leq j, \\
            ~~ 0 ~~~~~ \textrm{if} ~~ s_q < \delta. \\
        \end{cases}
    \end{split}
\end{equation}


\paragraph{An alternative adapter} We can replace LoRA with K-adapter~\cite{wang2020k}.
In K-adapter, $f$ is a transformer layer (denoted as $\textrm{TRM}(x)$), and $g$ is multiple transformer layers with two projection layers (denoted as $\textrm{KIA}(x)$)).
That is, $f(x)=\textrm{TRM}(x)$, consisting of one self-attention \& two feed-forward layers.
In the original paper~\cite{wang2020k}, $g(x)$ consists of multiple transformer layers and up\&down projection layers.
For K-adapter, we set a simple version with only a single transformer layer,
as follows:
\begin{equation}
    \label{eq_lora2}
    h = \textrm{TRM}(x) + \textrm{KIA}(x)
\end{equation}
where the parameters in $\textrm{TRM}$ are fixed and that in $\textrm{KIA}$ is trainable on target knowledge.

\section{Experiment}

In this section, we demonstrate the effectiveness of our approach on \task.



\paragraph{Datasets}
We evaluate our method on the following closed-book QA datasets: 

(1) \textbf{Zero-shot Relation Extraction (zsRE)}: \citet{levy2017zero} build relation-specific QA pairs, and \citet{de2021editing} utilize this dataset for a closed-book QA task.
This set provides question paraphrases based on the same fact and answer.
We split this set into two groups ($\mathcal{K}_s$ and $\mathcal{K}_t$) that do not share the same facts. 
To validate generalization, we build held-out sets ($\mathcal{P}_s$ and $\mathcal{P}_t$) that are not used in training process.
For this, we sample one QA pair among paraphrases based the same fact as $\mathcal{P}$.

(2) \textbf{Natural Questions (NQ)} + \textbf{SituatedQA}:
\citet{kwiatkowski2019natural} build NQ -- a large-scale QA dataset based on user queries. 
We consider NQ as source knowledge $\mathcal{K}_s$ except outdated facts based on SituatedQA.
\citet{zhang2021situatedqa} proposed SituatedQA identifying temporal- and geographical-dependent questions on a subset of NQ.
We use the temporal-dependent QA pairs as $\mathcal{K}_t$, which are annotated based on 2021 dump of Wikipedia.
For $\mathcal{P}_s$ and $\mathcal{P}_t$, as both NQ and SituatedQA do not provide paraphrases, we follow \cite{de2021editing} using back-translation for generating paraphrases.
    

\paragraph{Implementation}
For T5 model, we use a large version with total 770M parameters.
In our experiment, we assume that the old model $\theta^{old}$ storing source knowledge is available.
For NQ, we used the open-source pre-trained model\footnote{https://huggingface.co/google/t5-large-ssm-nq} as the model $\theta^{old}$.
For zsRE, we load and train T5 model\footnote{https://huggingface.co/google/t5-large-ssm} on source knowledge.
For training, we set batch size 64 on 4 RTX3090 GPUs,
and used Adam~\cite{KingmaB14} optimizer with learning rate 4e-4.
For development set, we sample each 1K from $\mathcal{K}_s$, $\mathcal{K}_t$, and select the maximum harmonic mean of their accuracies as a best model.
As a hyper-parameter,  we search $\delta$ in a range of [0,1] with 0.05 step size, and found the best value ($\delta$=0.9) based on development set. 
As embedding memory $\mathcal{M}$, we used additional parameters: 60M for zsRE and 8.5M for NQ. The size of the memories can be reduced by several techniques, such as random projection~\cite{luan2020sparse} and binary encoding~\cite{yamada2021efficient}, which is left out of our focus.

\paragraph{Comparison with baselines} 
We compare our method with baselines, as mentioned in Section 3.2; Fine-tuning (B-\RNum{1}), RecAdam (B-\RNum{2}), LoRA (B-\RNum{3}), and K-adapter (B-\RNum{4}).
When re-implementing K-adapter, we do not freeze the parameters of decoder, unlike in the original paper~\cite{wang2020k}, because the performance is not changing when freezing.
We train each model until 80 epochs and select a best model by the harmonic mean of source/target knowledge in development set.

\begin{table}[t]
\centering
\scalebox{0.88}{
\begin{tabular}{lcccc}
\noalign{\hrule height 1pt}
& \multicolumn{4}{c}{The total \# of examples}  \\ 
\cmidrule(lr){2-5}
& $\mathcal{K}_s$ & $\mathcal{P}_s$  & $\mathcal{K}_t$ & $\mathcal{P}_t$ \\ 
\noalign{\hrule height 1pt} 
zsRE (Large) & 60K & 24K & 60K & 24K \\ 
zsRE (Medium) & 60K & 24K & 30K & 12K  \\ 
zsRE (Small) & 60K & 24K & 15K & 6K  \\ \hline
NQ + SituatedQA & 59K & 32K & 8.3K & 1.6K \\ 
\noalign{\hrule height 1pt}
\end{tabular}
}
\caption{Statistics of datasets.}
\label{tab:dataset}
\end{table}

\begin{table*}[t]
\centering
\scalebox{0.92}{
\def\arraystretch{1.2}
\begin{tabular}{lccccccccccc}
\noalign{\hrule height 1pt} 
& & \multicolumn{5}{c}{zsRE Question Answering}  & \multicolumn{5}{c}{NQ (with SituatedQA)}  \\ 
\cmidrule(lr){3-7} \cmidrule(lr){8-12}
~~~~~ Method & \makecell{\# of Prams\\(train/total)} & $\mathcal{K}_s$ & $\mathcal{P}_s$ & $\mathcal{K}_t$ & $\mathcal{P}_t$ &  \makecell{F\slash U \\ Ratio}& $\mathcal{K}_s$ & $\mathcal{P}_s$  & $\mathcal{K}_t$ & $\mathcal{P}_t$ &  \makecell{F\slash U \\ Ratio} \\ 
\noalign{\hrule height 1pt} 
Model $\theta^{old}$ & - & 95.6 & 95.2 & 25.7 & 28.5 & -  & 96.6 & 94.9 & 35.3 & 33.7 & - \\ \hline
B-\RNum{1}: ~~~ Fine-tuning & 737M / 737M & 76.7 & 70.6  & 92.6 & 85.9 & 0.284 & 92.9 & 82.5 & 94.9   & \textbf{92.9} & 0.435 \\
B-\RNum{2}: ~~ RecAdam  & 737M / 737M & 80.5 & 74.7 & 91.6 & 83.5 & 0.230 & 93.1 & 82.1 & 93.8 & 92.1 & 0.419 \\
B-\RNum{3}:  ~ K-adapter & 538M / 840M & 80.5 & 70.8  & \textbf{96.4}  & 89.6  &  0.215 & 94.4 & 81.4 & 94.8  & 89.4  & 0.259 \\
B-\RNum{4}: ~ LoRA & ~ 62M / 799M & 71.1 & 62.9  & 92.9 & 84.8 & 0.366  & 89.8 & 74.0 & 94.0  & 90.5    & 0.800  \\
Ours (+K-adapter) & 538M / 840M & 86.3 & 78.9  & \textbf{96.4}  & \textbf{91.1} & 0.132  & \textbf{95.6} & 88.1  & 94.9 & 90.3 & 0.118\\
Ours (+LoRA) & ~ 62M / 799M & \textbf{90.5} & \textbf{90.6}  & 95.3 & 89.4 & \textbf{0.073} & \textbf{95.6} & \textbf{95.2}  &  \textbf{95.1} & 90.0 & \textbf{0.117} \\

\noalign{\hrule height 1pt} 
\end{tabular}}
\caption{The comparison of the continual learning results  on zsRE (Large) and NQ datasets. We measure the accuracies on the knowledge $\mathcal{K}_s, \mathcal{K}_t$, and the paraphrase knowledge $\mathcal{P}_s, \mathcal{P}_t$, with the F/U ratio. }
\label{main_result}
\end{table*}

\begin{figure*}[t]
	\centering
	\includegraphics[width=160mm]{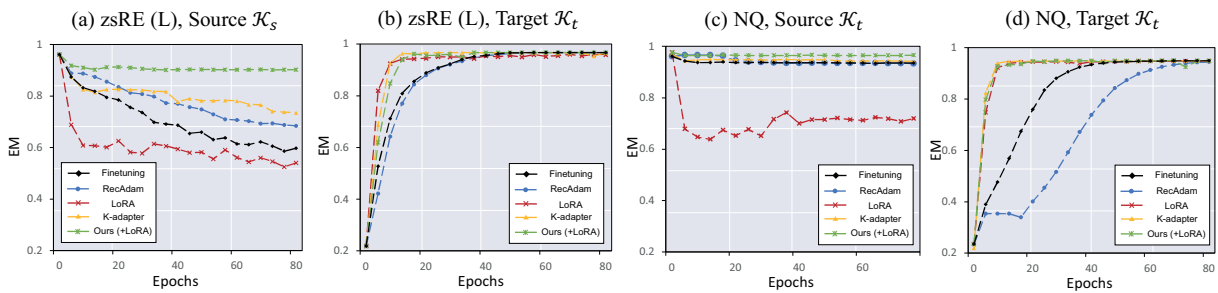}
	\caption{Accuracies of ours and baselines over training epochs.}
	\label{figure3}
\end{figure*}

\subsection{R1: Comparing Ours with Baselines}

Table \ref{main_result} shows our main experimental results on two CBQA datasets.
First, the model $\theta^{old}$ memorizes the source knowledge $\mathcal{K}_s$ well and generalizes on the paraphrase set $\mathcal{P}_s$ as well, showing high accuracy on both datasets.
After training on $\mathcal{K}_t$, all models perform well on $\mathcal{K}_t$ and $\mathcal{P}_t$.
These results indicate that these models are at least appropriate for memorizing training data in the current task.

Meanwhile, while acquiring $\mathcal{K}_t$, the models show variant results on $\mathcal{K}_s$ and $\mathcal{P}_s$, which have the different ability of retaining previous knowledge against forgetting.
In Fine-tuning (B-\RNum{1}), its performances on source knowledge $\mathcal{K}_s$ and $\mathcal{P}_s$ decrease as training epochs (see Figure~\ref{figure3}).
RecAdam (B-\RNum{2}) alleviates the forgetting problem of fine-tuning, but the performance gains are marginal on two datasets.
K-adapter (B-\RNum{3}) shows the strong performance on $\mathcal{K}_s$ with less forgetting, however, does not perform well on $\mathcal{P}_s$ and $\mathcal{P}_t$ showing low generalization.
Because LoRA (B-\RNum{4}) has the fewest trainable parameters, 
its forgetting is more aggravated, showing the worst performance on $\mathcal{K}_s$ and $\mathcal{P}_s$ in both zsRE and NQ.
Ours with either K-adapter or LoRA shows the best performance on $\mathcal{K}_s$ and $\mathcal{K}_t$.
In terms of the F$\slash$U ratio, our method also shows the lowest loss when updating one new example.
Figure \ref{figure3} shows how the performance of each model changes over training epochs, on the development set.

\paragraph{Ablation study}
In an ablation study, we test which component has the higher impact on memorizing implicit knowledge, on paraphrase set $\mathcal{P}_s$ and $\mathcal{P}_t$.
In our method with LoRA, the function $f$ in Eq. (\ref{aggregation}) can be applied to any projection layer in transformers.
While the original work~\cite{hu2021lora} applies to query- and value-matrices ($W_Q, W_V$) in self-attention, we consider feed-forward layers ($W_{FF}$), as well as self-attention.
In addition, we observe how does the performance vary when the number of parameters increases by controlling rank $r$.
In Table~\ref{ablation}, we empirically found applying feed-forward layers is more effective than query and value projection, especially on target knowledge $\mathcal{P}_t$.
These results indicate that memorizing factual knowledge is more relevant with a feed-forward module, which is consistent with the views in \cite{sukhbaatar2019augmenting,geva2020transformer}.


\begin{table}[t]
\centering
\scalebox{0.8}{
\def\arraystretch{1.2}
\begin{tabular}{c|ccc|ccc|c}
\noalign{\hrule height 1pt} 
Type & \multicolumn{3}{c|}{$W_Q$,$W_V$} & \multicolumn{3}{c|}{$W_{FF}$} & All \\
Rank $r$  & 16 & 64 & 256 & 16 & 64 & 256 & 256 \\ \hline
$\mathcal{P}_s$ &94.9 &  95.2 &  95.5 &  95.1 &  95.0 & 95.2 & 95.2 \\
$\mathcal{P}_t$ & 59.6 & 65.1 & 65.5 & 87.1 & 89.2 & 90.0 & 89.3 \\

\noalign{\hrule height 1pt} 
\end{tabular}}
\caption{An ablation study}
\label{ablation}
\end{table}

\begin{figure*}[t]
	\centering
	\includegraphics[width=161mm]{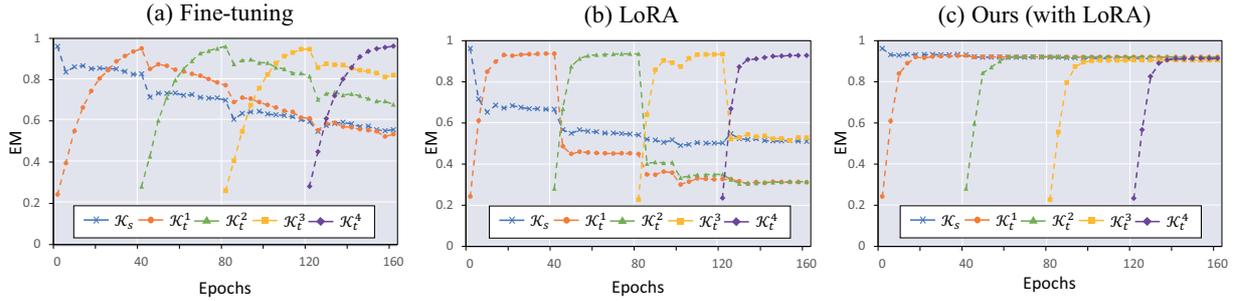}
	\caption{The accuracies on multiple knowledge sources (K=5) over training epochs for zsRE.}
	\label{figure4}
\end{figure*}

\begin{figure}[ht]
	\centering
	\includegraphics[width=78mm]{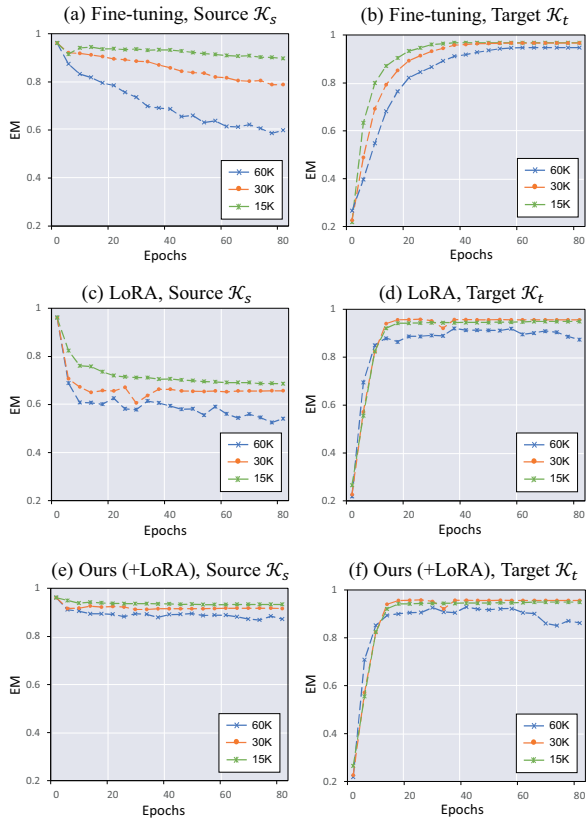}
	\caption{Accuracies over varying size of zsRE.}
	\label{figure5}
\end{figure}

\subsection{R2: Accumulating over Multiple $\mathcal{K}_t$}

To evaluate the scalability of our method on multiple $\mathcal{K}_t$ ($>$2),
we assume multiple updates (five-phase) with smaller amount of examples, by splitting target knowledge $\mathcal{K}_t$ in zsRE (Large, 60K), into four sets, from $\mathcal{K}_t^1$ to $\mathcal{K}_t^4$ (each 15K).
In this experiment, we train models during 40 epochs/phase.
To generalize for LoRA baseline, we aggregate multiple $g_k$ by addition, by activating all the switches at inference, \ie, $\sigma_{1:M}(x)=1$ in Eq. (\ref{multiple_switch}).
This setting assumes that this baseline cannot leverage our selector to organize the storage of implicit knowledge.
Figure \ref{figure4} shows the performances of Fine-tuning, LoRA, and Ours, over training epochs.
In fine-tuning, the accuracy on source knowledge keeps dropping during the whole training process.
In LoRA, multiple updating deteriorates memorizing target knowledge stored in adapters, faster than source knowledge stored in the original parameters.
This indicates that the fewer parameters, the faster the forgetting.
In contrast, our method consistently outperforms the baselines, by retaining five knowledge, with forgetting less.
To summarize these results, sequential updates aggravate forgetting of the fine-tuning method, which can be overcome through the selective use of adapters.

\subsection{R3: Over varying Size of $\mathcal{K}_t$}

As the size of target knowledge increases, it makes LMs suffer from more forgetting, increasing the distance between before- and after-parameters.
In this section, we observe how does the performance of each model vary as different sizes of $\mathcal{K}_t$.
Figure \ref{figure5} shows the the accuracies of zsRE datasets (Large-60K, Medium-30K, Small-15K), over training epochs.
On source knowledge $\mathcal{K}_s$, the performance of fine-tuning and LoRA keeps dropping, and the accuracy drops are proportional to the size of target knowledge.
Meanwhile, our method with LoRA consistently maintains high performance, which is not sensitive to training epochs.
On target knowledge $\mathcal{K}_t$, the performances of three models reach high accuracy.
However, our method on Large zsRE shows unstable performance at the end of training, which may need to use early stopping.

\begin{table}[t]
\centering
\def\arraystretch{1.8}
\scalebox{0.82}{
\begin{tabular}{|cc|cc|}

    \hline
    \multicolumn{2}{|c|}{}  & \multicolumn{2}{c|}{Ground-truth}  \\ \cline{3-4} 
    \multicolumn{2}{|c|}{}   & \multicolumn{1}{c|}{Source}  &  Target  \\ \hline
    \multicolumn{1}{|c|}{\multirow{2}{*}{\makecell{Selector \\ Prediction}}}
    & Source & \multicolumn{1}{c|}{\cellcolor{red!55}\makecell{~~19527~~\\(40.7\%)}} &  \cellcolor{red!12}\makecell{~~~~854~~~~\\(1.8\%)}   \\ \cline{2-4} 
    \multicolumn{1}{|c|}{}  & Target & \multicolumn{1}{c|}{\cellcolor{red!28}\makecell{~~~4473~~~\\(9.3\%)}}   & \cellcolor{red!75}\makecell{23146\\(48.2\%)}  \\ \hline

\end{tabular}}
\caption{The confusion matrix of Selector.}
\label{table6}
\end{table}

\begin{table}[t]
\centering
\def\arraystretch{1.8}
\scalebox{0.78}{
\begin{tabular}{|cc|cc|}
    \hline
    &        & \multicolumn{2}{c|}{Ground-truth}                \\ \cline{3-4} 
    &        & \multicolumn{1}{c|}{Source}       & Target       \\ \hline
    \multicolumn{1}{|c|}{\multirow{2}{*}{\makecell{Selector \\ Prediction}}}
    & Source & \multicolumn{1}{c|}{ 95.3 } & 35.1    \\ \cline{2-4} 
    \multicolumn{1}{|c|}{}  & Target & \multicolumn{1}{c|}{70.8 (0.0)}   & 91.7 (97.4)  \\ \hline
\end{tabular}}
\caption{The accuracies of Ours/Retrieval in four cases.}
\label{table7}
\end{table}

\subsection{Analysis of Selector}
In Table \ref{table6}, we show the distribution of selector's predictions and the ground-truths, in our experiment on zsRE (Large).
Nearest Neighbor-based selector successfully classifies 88.9\% of examples, while 11.1\% failed.
In our method, if the selector classifies an input as target knowledge, the plug-in $g$ is activated.
Instead of the use of $g$, we can retrieve answers aligned with questions in $\mathcal{M}$, not generate them.
We compare our generation with the retrieval in each case of Table \ref{table6}.
Table \ref{table7} shows the accuracy of predicting the answers,
where the numbers in each cell indicate EM of our generation (retrieval: in parentheses).
If an example in source knowledge is incorrectly classified as target, there is no relevant fact in $\mathcal{M}$, thus the accuracy in this case is zero.
In contrast to Retrieval, our generative method is robust in this case, achieving 70.8\% EM, because ours with $g$ learned the source knowledge.

\section{Conclusion}
This paper studies how to accumulate new knowledge to LMs that stores existing knowledge.
We propose a simple yet effective method to update target knowledge into new parameters, preventing from forgetting source knowledge.
On two datasets: zsRE and NQ, our empirical results show that our proposed method can improve existing approaches for continual learning or task adaptation.

\section{Acknowledgement}
This research was supported by SNU-NAVER Hyperscale AI Center, and IITP grants funded by the Korea government
(MSIT) [2021-0-02068 SNU AIHub, IITP-2022-2020-0-01789].

\bibliography{anthology}
\bibliographystyle{acl_natbib}

\end{document}